\documentclass{article}

\usepackage{arxiv}

\usepackage[utf8]{inputenc} % allow utf-8 input
\usepackage[T1]{fontenc}    % use 8-bit T1 fonts
\usepackage{hyperref}       % hyperlinks
\usepackage{url}            % simple URL typesetting
\usepackage{booktabs}       % professional-quality tables
\usepackage{amsfonts}       % blackboard math symbols
\usepackage{cases}
\usepackage{nicefrac}       % compact symbols for 1/2, etc.
\usepackage{microtype}      % microtypography
\usepackage{lipsum}
\usepackage{graphicx}

\title{Reduced Focal Loss: \emph{1st} Place Solution to xView object detection in Satellite Imagery }

\author{
  Nikolay A. Sergievskiy \\
  XIX.ai Inc.\\
  \texttt{dereyly@gmail.com} \\
   \And
 Alexander A. Ponamarev \\
  IQVIA \\
  \texttt{alex.ponamaryov@gmail.com} \\
}

\begin{document}
\maketitle

\begin{abstract}
This paper describes our approach to the DIUx xView 2018 Detection Challenge [1]. This challenge focuses on a new satellite imagery dataset. The dataset contains 60 object classes that are highly imbalanced. Due to the imbalanced nature of the dataset, the training process becomes significantly more challenging. To address this problem, we introduce a novel Reduced Focal Loss function, which brought us 1st place in the DIUx xView 2018 Detection Challenge. 

\end{abstract}

% keywords can be removed

\section{Introduction}
DIUx xView 2018 Detection Challenge\cite{lam2018xview} is currently the largest publicly available object detection dataset of satellite objects. The dataset contains approximately 1 million objects, which are represented by 60 object classes. These satellite imageries within the dataset are obtained with an accuracy of 0.3 meter. The total area covered by the marked-up data amounts to 1,400 km2. The object detecting task may be compared to COCO [5], but this competition has its own specific parameters. For one, the imageries are unique satellite images, which present an array of small objects . Each object does not greatly change its scale, unlike objects in COCO, and the objects can rotate 360 degrees. In addition, the imageries possess challenging and  indistinguishable features and the data has an inherently imbalanced nature. For example, classes such as Small Cars and Buildings occur 200-300K times within the dataset, while Railway Vehicles and Towers each have 100 instances. 
\begin{figure}
  \centering
  \includegraphics[height=9cm]{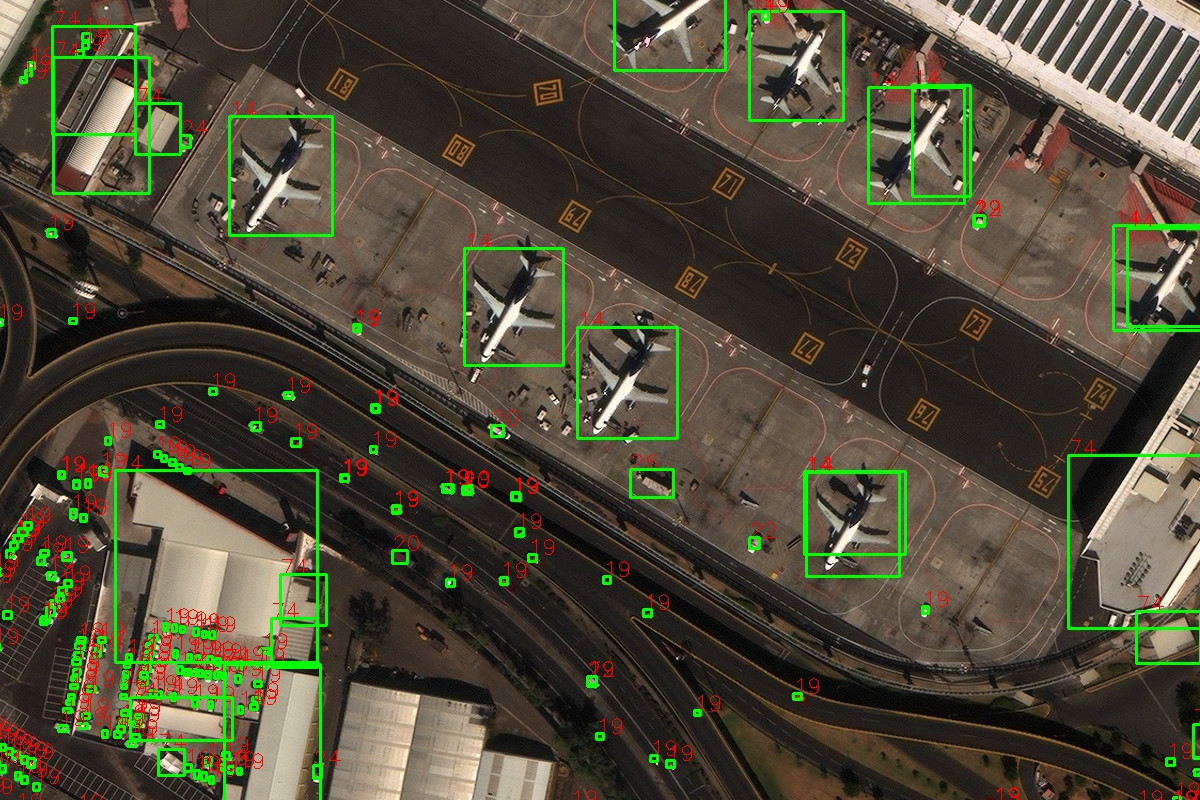}
  \caption{Detection visualisation on xView dataset}
  \label{fig:fig1}
\end{figure}

\section{Method}
\label{sec:headings}
\subsection{2.1 Related work}
Each image is passed through a backbone architecture, such as resnet, and a feature map is taken after each block. RPN is applied on each feature map, but shares the same weights across all levels, which produces object proposals. The best proposals are refined using two fully connected layers: Feature Pyramid Network (FPN) [2] first collects predictions from all levels, then applies NMS to them. It then selects the best level to which RoIAlign is applied, based on the size of the proposal. The results of the base FPN detector can be found in Table\ref{table:1}
\begin{table}[h!]
\centering
\begin{tabular}{ |c|c|c|c| } 
 \hline
Model name & mAP & Recall & mRecall \\
 \hline
baseline (SSD) & 21.78* & & \\
FPN-50 & 24.37 & 0.752 & 0.672 \\
FPN-50 additional anchors & 25.62 & 0.774 & 0.685 \\
Retina Net & 9.7 & & \\
FPN-50 Focal loss & 19.01 & 0.48 & 0.66 \\
FPN-50 Reduced Focal loss & 26.92 & 0.668 & 0.741 \\
FPN-50 IoU Reduced Focal loss & 27.44 & 0.687 & 0.748 \\
FPN-50 IoU Reduced Focal loss random skip most frequent & 28.32 & 0.612 & 0.775 \\
 \hline
\end{tabular}
\caption{The results of different models on the validation dataset. The second column is a mean average precision on the validation dataset (the score on public LB is 1-2\% lower then on the validation). The baseline SSD is a model provided by the competition organizers, and accuracy is indicated on public LB. FPN-50 experiments use the ResNet-50 and FPN architecture, and a two-stage detector Faster RCNN. The third row is standard recall after the RPN stage. the fourth row is a mean per class recall after RPN, which we called mRecall.
}
\label{table:1}
\end{table}

\subsection{Reduced Focal Loss}
Imbalanced data presents a significant challenge for training machine learning models. Various methods were used to counteract this problem, such as re-sampling and Synthetic Minority Over-sampling Technique (SMOT) \cite{chawla2002smote}. In this paper, we propose a novel Reduced Focal loss, a method which is designed to improve training process on original imbalanced data.

Focal Loss \cite{lin2017focal} is an effective strategy for training object detection models that has an inherent imbalance of classes. Focal Loss can be expressed as follows:
\begin{equation}
FL(pt)=-(1-pt)^{\gamma}\log{pt}
\end{equation}
Where $(1-pt)^{\gamma}$ scales the loss function, and pt is the probability of the ground truth class
The first approach we used involved applying the concept of the focal loss to two stages of the faster RCNN framework by replacing original loss functions with their focal loss equivalents. As such, a binary loss applied to RPN head was swapped for a version of the focal loss. Additionally, cross entropy in the Faster RCNN head was also replaced with focal loss. Unfortunately, this approach did not improve our results.

Focal Loss applies exponentially higher weights to hard samples, which helps to reduce the loss contribution of well classified samples. However, these exponentially higher weights lead to an extreme effect of hard and mislabeled samples (as in many other datasets, XView contains some mislabeled data). In order to better understand the behavior of Focal Loss, we looked at the recall of RPN. We noticed that the recall produced by Focal Loss was much worse than its binary loss counterpart. This observation led to our hypothesis that the shift of the focus towards high loss, produced by Focal Loss, results in a more complex model. This model passes a lower number of good proposals to the next stage (Fast RCNN), which results in lower recall. \\
\textit{It is important to note that Focal Loss minimizes the contribution of well classified samples and drives the focus towards hard samples. Put differently, Focal Loss is a soft approach to hard example mining. } \\
This behavior contradicts the two-stage approach of Faster RCNN \cite{ren2015faster} where RPN seeks to maximize recall, (allowing false positives), while the next stage, Fast RCNN, has the capacity to classify proposals correctly.

After empirically proving that the drop in recall was caused by Focal Loss, we proposed a modified version of this function that we named Reduced Focal Loss. Reduced Focal Loss has two goals - 1) to soften the response of the loss function to hard samples, and  2) continuing to perform hard example mining. To soften the response of the loss function to hard samples, we applied “flat” weights to positive samples with probabilities less than a particular threshold. In order to continue minimizing the impact of well classified samples, we retained the Focal Loss approach scaled to reflect the threshold.

\begin{figure}
  \centering
  \includegraphics[height=9cm]{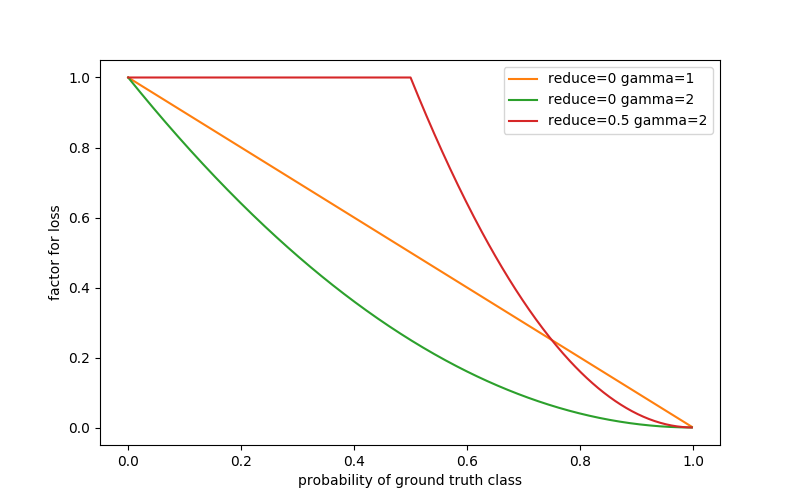}
  \caption{Reduced Focal Loss - Cut-off Factor.}
  \label{fig:fig2}
\end{figure}

Reduced Focal Loss can be described as follows:
\begin{equation}
    RFL(pt)=- fr(pt, th)\log{pt}
\end{equation}
Where $fr(pt, th)$ is a cut-off factor that scales loss function according to the following formula:

\begin{numcases}{f(x)=}
  1 & : $pt < th$ \nonumber \\
  \frac{(1-pt)^{\gamma}}{th^{\gamma}} & : $pt \geq th$
\end{numcases}
The cut-off factor (illustrated in fig\ref{fig:fig2}) solves the problem of a drop in the recall of RPN (table \ref{table:1}. Column 2 and 3). As compared with Cross Entropy, Reduced Focal Loss delivers marginally lower total recall. However, Reduced Focal Loss dramatically improves the recall of rare classes, which addresses the problem of imbalanced classes.
Reduced Focal Loss also improves the results of Fast RCNN performance. Reduced Focal Loss produces the same loss as Cross Entropy function in the high loss zone (where the high loss zone defined as the probability of a positive class in the 0.0 to 0.5 range). A built-in switch between Focal Loss and Cross Entropy integrated into Reduced Focal Loss helps to mitigate the extreme impact of rare classes and mislabeled data, while still reducing the contribution of well classified samples.

\subsection{Random undersampling of most frequent classes}
Reduced Focal Loss helps to improve the overall performance of the model. However, the comparison between the model trained without two prevalent classes, Small Cars and Buildings, and the model trained on all classes shows that Reduced Focal Loss does not fully remove the problem of imbalanced classes. The model trained with Reduce Focal Loss on a modified dataset that excludes Small Cars and Buildings demonstrated marked improvement in mAP for rare classes. This improvement in performance led to the conclusion that the model training can be further enhanced by reducing class imbalance.

In order to reduce the influence of prevalent classes, we utilized random undersampling, which is designed to remove prevalent classes from the training with a certain probability (training parameter). This approach allowed us to improve mAP for rare classes, using a model trained end-to-end. The resulting end-to-end model outperformed a model, which was trained on the modified dataset (without Small Cars and Buildings entirely). However, this approach lessened mAP for Small Cars, while keeping mAP for Buildings more or less flat (in comparison with a model trained on all classes without undersampling (appendix.Table 2)).

\section{Experiments}
We present experimental results on object detection in the context of overhead imagery of the DIUx xView 2018 Detection Challenge where our approach won first place. The DIUx xView 2018 Detection Challenge is based on the xView Dataset that consists of 846 annotated images. For training purposes, we split the xView Dataset into training and validation datasets containing 742 and 104 images respectively.

The training dataset was further cropped into smaller images (700x700 resolution) with an overlap equal to 80 pixels. Each of the resulting images was further augmented with the following rotations: 10, 90, 180, 270 degrees. As a result, the size of the final training dataset increased to 63,535 images. The aforementioned approach was proven to be effective for the training of machine learning models on satellite images. In addition to the image rotation, we applied the following online augmentation techniques: random flip, color jittering, and scale jittering (ranging between 500 to 900 pixels).

The code was implemented using the Pytorch version of Detectron package \cite{detectron_pytorch}. For the object detection we used FPN Faster RCNN framework with two fully connected layers in heads, and Resnet-50 as a backbone. As compared to the default settings, we increased the batch size on the heads to 1024 and RPN batch size to 512. Other settings of the training pipeline include:
\begin{itemize}
\item foreground to background ratio - 0.5
\item probability threshold for Reduced Focal Loss of RPN - 0.5
\item probability threshold for Reduced Focal Loss of Fast RCNN - 0.25
\end{itemize}
The final challenge winning model was trained on a rig that includes two NVIDIA 1080Ti GPU’s. The training was done in two stages: an initial training and a “fine tuning”. The initial training was done for 24 hours (180 thousand iterations) with the learning rate scheduled as follows:
\begin{itemize}
\item first 120 thousand iterations - 0.005
\item 120 - 140 thousand iterations - 0.0005
\item the rest of the training - 0.00005
\end{itemize}
The fine tuning stage was done on a combined dataset comprised of training and validation data with 0.001 learning rate.
With these settings, the main model of Reduced Focal Loss was trained, for which the mAP is on validation samples 28.32 and 26.75 on Public LB.
The final decision is based on an ensemble of models with test-time augmentation:
\begin{enumerate}
\item scale: 1.2, model: “Additional anchors”, rotation 90
\item scale: 1, model: “Reduced Focal Loss”
\item scale: 0.8, model: “Reduced Focal Loss”, rotation 90 
\item scale: 1.2, model: “ Baseline (SSD)”
\item scale: 1, model: “ Baseline (SSD)”
\item scale: 0.7, model: “ Baseline (SSD)”
\item scale: 0.6, model: “ Baseline (SSD)”
\end{enumerate}
The models were combined using a modified voting method. The combined model won first place with a strong margin. The model achieved 31.74 on a public LB and 29.32 on Private LB.

\bibliographystyle{ieeetr}   %unstr
\bibliography{references.bib}


% George Kour and Raid Saabne.
% \newblock Real-time segmentation of on-line handwritten arabic script.
% \newblock In {\em Frontiers in Handwriting Recognition (ICFHR), 2014 14th
%   International Conference on}, pages 417--422. IEEE, 2014.

% George Kour and Raid Saabne.
% \newblock Fast classification of handwritten on-line arabic characters.
% \newblock In {\em Soft Computing and Pattern Recognition (SoCPaR), 2014 6th
%   International Conference of}, pages 312--318. IEEE, 2014.

% \bibitem{hadash2018estimate}
% Guy Hadash, Einat Kermany, Boaz Carmeli, Ofer Lavi, George Kour, and Alon
% \newblock Estimate and replace: A novel approach to integrating deep neural
%   networks with existing applications.
% \newblock {\em arXiv preprint arXiv:1804.09028}, 2018.
%\end{thebibliography}

\end{document}